\documentclass[letterpaper]{article}

\usepackage{natbib,alifeconf}  

\title{ALIFE2022 template}

\title{Amorphous Fortress: \\Observing Emergent Behavior in Multi-Agent FSMs}
\author{M Charity$^{\dag}$, Dipika Rajesh$^{\ddag}$, Sam Earle$^{\dag}$, Julian Togelius$^{\dag}$ \\
\mbox{} \\
    $^{\dag}$ New York University, New York, NY 11220\\
    $^{\ddag}$ Independent \\
    mlc761@nyu.edu, dipika.rajesh@gmail.com, se2161@nyu.edu, julian@togelius.com
} 

\usepackage{todonotes}
\usepackage{multirow}
\usepackage{caption}
\usepackage{subcaption}
\usepackage{colortbl}
\usepackage{graphicx}
\usepackage[hidelinks]{hyperref}
\usepackage[multiple]{footmisc}

\usepackage{algpseudocode}
\usepackage{xcolor}
\usepackage[linesnumbered,ruled,vlined]{algorithm2e}

\SetCommentSty{mycommfont}

\SetKwInput{KwInput}{Input}                
\SetKwInput{KwOutput}{Output}              

\begin{document}
\maketitle

\begin{abstract}

We introduce a system called Amorphous Fortress---an abstract, yet spatial, open-ended artificial life simulation. In this environment, the agents are represented as finite-state machines (FSMs) which allow for multi-agent interaction within a constrained space. These agents are created by randomly generating and evolving the FSMs; sampling from pre-defined states and transitions. This environment was designed to explore the emergent AI behaviors found implicitly in simulation games such as Dwarf Fortress or The Sims. We apply the hill-climber evolutionary search algorithm to this environment to explore the various levels of depth and interaction from the generated FSMs


\end{abstract}

\section{Introduction}

Accidental cat poisonings in Dwarf Fortress, characters from the Sims choosing to ``woo-hoo'' the Grim Reaper, using bullet-time and a bomb to literally fly across the map in Breath of the Wild---these are just a few examples of unintentional behaviors in games. These emergent behaviors can lead to very memorable but sometimes irreplicable experiences for the player---with the potential to become something like rumors or urban legends for the game's community. 
 Our goal is to use evolution to discover open-ended environments where this kind of unexpected gameplay is likely emerge.

Previous works in both research and game design have explored both simulations and emergent AIs in their own unique systems. Some examples include the computer simulation games Tierra\footnote{http://tomray.me/tierra/index.html}, Avida\footnote{https://github.com/devosoft/avida}, and Conway's Game of Life \citep{gardner1970mathematical}, each with diverse and stimulating generated narratives that stem from these AI interactions. These interactions were never originally intended or perceived of by their human designers, but nevertheless, create evocative situations and experiences within the system. Exploring these artificial experiences could allow AI designers to find interesting agent behaviors naturally without having to hard-code these behaviors within the initial system.


We introduce an open-ended simulation for evolving artificial life with abstract and emergent behaviors. This system will allow us to observe the interactions that emerge in multi-agent simulations and how innovative and interesting behaviors can be generated from a small pre-defined set of primitive actions and conditions in a confined space. We conduct a hill-climbing evolutionary experiment to examine and discuss some emergent scenarios and entities with these evolved FSMs.

\begin{figure*}[ht!]
    \centering
    \includegraphics[width=0.98\textwidth]{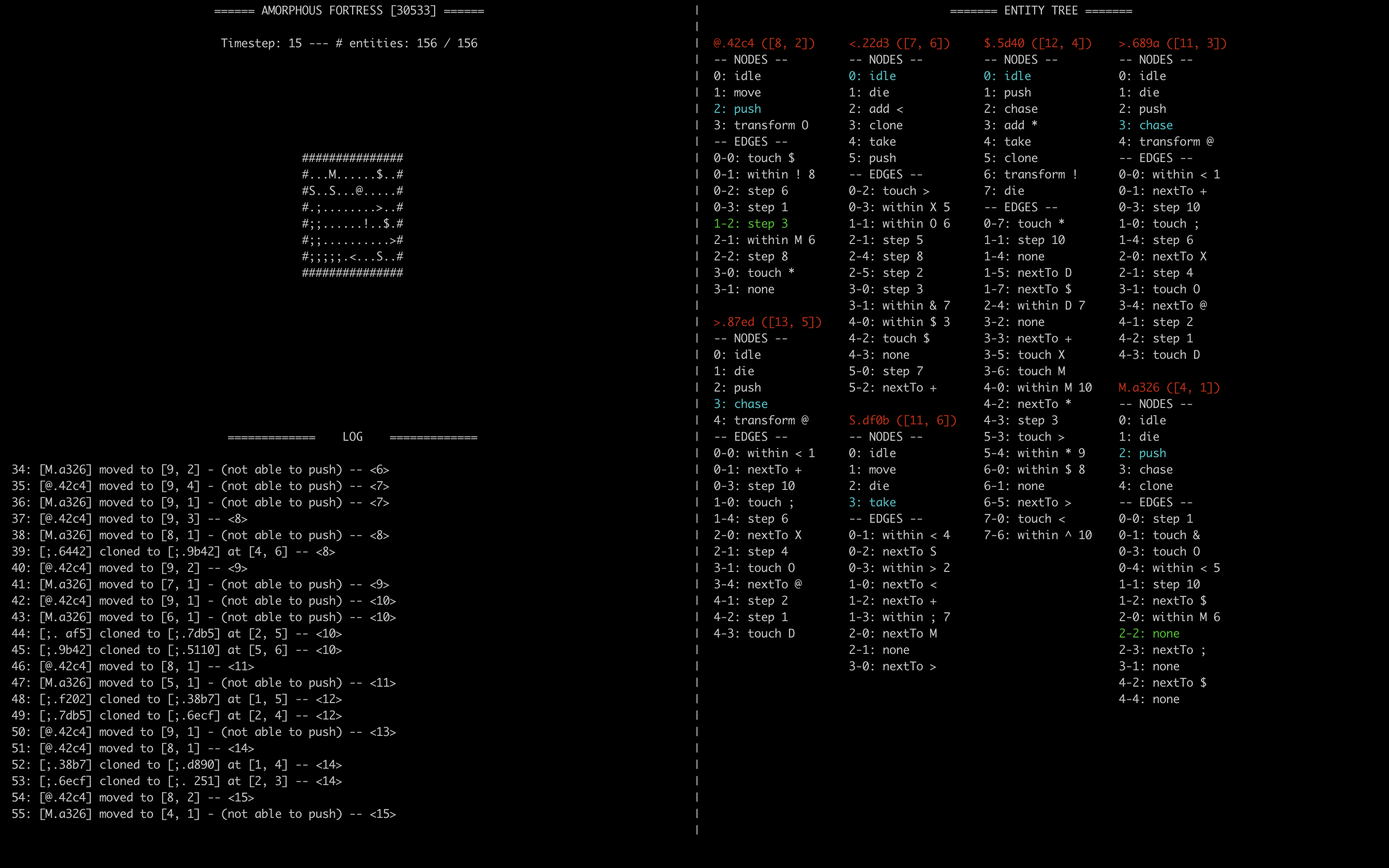}
    \caption{Terminal Interface of the Amorphous Fortress. On the left side is the rendering of the entities with the logged output below. On the right side are the activated nodes and edges of the instance entities.}
    \label{fig:SimulationTerminal}
\end{figure*}

\section{Background}

This system combines many notable AI methodologies including rule generation for the different action nodes, finite state machines for the agent graph generations, and the ($1+1$) Evolutionary Search Algorithm for the agent diversity exploration. Other works involving simulation games and emergent AI also greatly influence this work.

\subsection{Rule Generation}

To create the range of behaviors for the artificial agents in the context of the game environment---we take inspiration from rule generation algorithms. As \citep{togelius2008experiment} demonstrates, games can be evolved by exploring and mutating their rules. Previous works such as ANGELINA~\citep{cook2016angelina} and Sturgeon-MK3~\citep{cooper2023sturgeon} also use game rule manipulation to create and recreate games and environments without having to directly modify the underlying source code of an engine or system. Board games also benefit from the automated creative process and can result in interesting and aesthetically satisfying games without any human intervention ~\citep{browne2012yavalath}. We explore the concept of rule generation to create new scenarios and multi-agent behaviors that do not require pre-defined human involvement with the Amorphous Fortress environment.


\subsection{Finite-State Machines}

Finite state machines (FSMs)---as defined by~\cite{georgios2018artificial}---are an abstract representation of an interconnected set of actions, states, and transitions represented by a graph. These graphs typically represent conditional relationships between nodes and actions. 
FSMs are one of the key components of classic game AI. They have been used  to create simple non-playable character behaviors in a very large number of games, including Pac-man, Half-life, and F.E.A.R \citep{orkin2006three}. Previous works have also explored the use of FSMs as solver agents in AI competitions such as Starcraft \citep{smolyakov2019design} and Pommerman \citep{zhou2018hybrid} and as agents in a mixed-initiative simulation system \citep{olsen2009beep}. 
Despite their popularity, vanilla finite state machines lack dynamicity compared to other more recent and sophisticated approaches to NPC AI. For example, Goal-Oriented Action Planning~(GOAP, \cite{orkin2006three}), popularized by its use in F.E.A.R., combines the procedural representation of FSMs with planning, allowing agents to search for optimal outcomes over an FSM defining lower-level behavioral relations. By allowing designers to declare goals for such planning agents, GOAP makes it easier to simulate intentional, long time-horizon behaviors in NPCs. 
Here, we limit ourselves in scope to straight-ahead FSMs, with single-behavior nodes, and posit that applying evolutionary search to such a representation---despite its simplicity---is enough to lead to an interesting diversity of NPC behavior profiles, leaving the evolution of more general-purpose NPC AI description languages to future work.






\subsection{Evolutionary Algorithms}


Evolutionary Algorithms, centered around the idea of ``survival of the fittest", iteratively mutate feature vectors (genomes) to create a population, score the population based on a fitness criterion, and select the best genomes for the next generation. Some examples of evolutionary algorithms include genetic algorithms, MAP-Elites, and (mu+lambda) evolution strategies. These algorithms have been used to generate game AI components such as dungeons \citep{baldwin2017mixed} and enemies \citep{pereira2021procedural}. The (1+1) evolution strategy, also known as the ``Hillclimber" algorithm~\citep{norvig2002modern}, is a greedy evolutionary algorithm that incrementally searches for the local optima to find the best genomes among a population. The algorithm continuously chooses the genome with the higher fitness score between the parent and the child of a given population to serve as the parent of the next population. We use the Hillclimber algorithm to explore the depths of the generated finite-state machine behaviors.

    

\subsection{Simulation games}

Simulating real-world events through games allows researchers to study and emulate phenomena in controlled environments while allowing for a variety of situations and interactions internal to the systems at play. Combined with artificial agents, these simulation games can explore how AI could interact with both the player and the world itself. The SimSim environment~\citep{charity2020say} evolves furniture arrangements in houses based on the Sims games to find novel designs while keeping the player alive and satisfied. Similarly, \citep{earle2020using} introduce a training environment for reinforcement learning agents in the game SimCity and examine population behaviors of cellular automata in Conway's Game of Life. \citep{green2021exploring} use a minimal clone of RollerCoaster Tycoon to generate a diversity of theme park layouts. Thus, simulation games provide a perfect testing ground for artificial agents to either engage with or generate artifacts that can be extrapolated to real-world scenarios.


\subsection{Emergent AI}

In both the games and game AI research fields, emergent AI has become an increasingly notable area of study. The breadth of interactivity afforded by non-playable characters and back-end artificially intelligent and adaptive systems allow for more individualized player experiences, some of which the game designers may have never intended. In games, many emergent phenomena result from NPCs interacting with each other based on some set of rules, such as in Dwarf Fortress, or from adapting to player behaviors and decisions, such as the AI director in Left 4 Dead. In games research, recent work has explored how artificial generators can appear ``haunted'' to a player when the generator creates highly abnormal samples \citep{kreminski2023generator} or when narratives emerge from level generations that stem from the individual player experience and interaction with the generated artifact \citep{nicholls2023darned}. The focus of this paper is to explore how emergent AI behavior from generated FSMs could inspire narratives or descriptions of entities that have emerged through abstract code, or how multi-agent systems could interact without any pre-made definition to create more complex situations not foreseen by the designer.

\section{Methods}

The Amorphous Fortress\footnote{We open-source our code here: \url{https://anonymous.4open.science/r/amorphous-fortress-BBDC/}} is an artificial life simulation system made up of a hierarchy of 3 components: entities (the agent class of the system) the fortress object (the environment class of the system) and the engine (the ``manager'' and main loop of the simulation). A configuration file can be provided upon initialization of the system to define a particular range of interactions and allow for the reproducibility of experiments with the framework. The following subsections describe each component in more detail and in the context of the experiments done for this paper.

\begin{figure}
    \centering
    \includegraphics[width=0.9\linewidth]{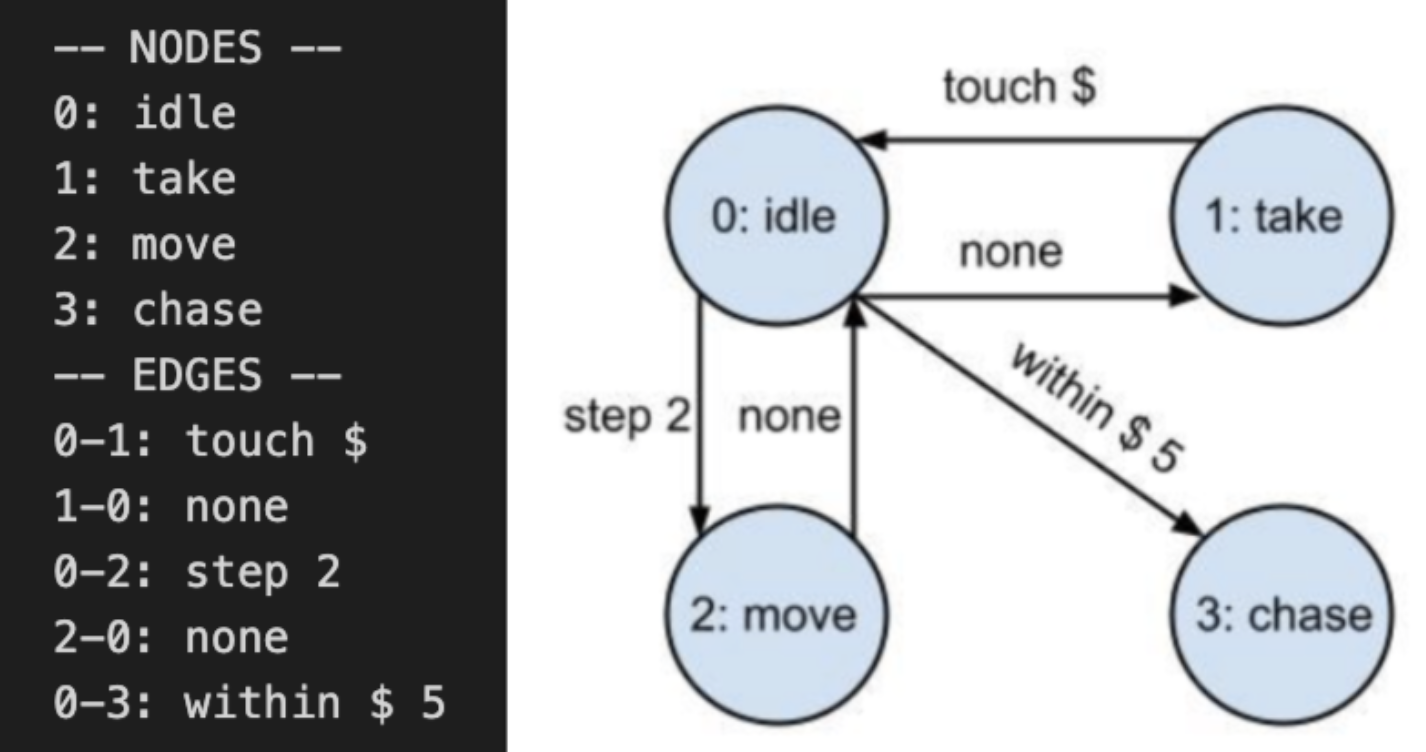}
    \caption{Example of both the text file output (left image) of the entity FSM representation and the translated visual graph representation (right image)}
    \label{fig:EntityFSM}
\end{figure}

\subsection{Entities}

Each ``entity'' of the Amorphous Fortress is defined by a singular ASCII character, a unique 4-bit identification hex number, a list of nodes, and a set of edges. 

The ASCII character-rendering was directly inspired by classic terminal-based rogue-likes such as Rogue and Dwarf Fortress and also allowed for more anthropomorphism with the objects and classes without explicit definitions of their identities. These characters can be any symbol that can be represented on a terminal but lack the color variance found in text-based games.

Because the system uses a finite-state machine as the basis of interaction and behavior, each node in the FSM graph represents a potential action state the entity object can be in. These actions are pre-programmed and define how an entity interacts with the environment. For this system, the possible action nodes are defined as follows: 
\begin{itemize}
    \item \textit{idle}: the entity remains stationary at the same position
    \item \textit{move}: the item moves in a random direction (north, south, east, or west) so long as there is no `wall' character at that space, as defined in the `fortress' entity
    \item \textit{die}: the entity character and any reference to it is removed from the fortress entirely
    \item \textit{clone}: the entity makes another instance object of its own class
    \item \textit{push}: the entity will attempt to move in a random direction but will push any object that is currently in that space to the next space over (if possible)
    \item \textit{take (char)}: the entity takes the targeted object that is represented by the specified character
    \item \textit{chase (char)}: the entity moves towards the position of the targeted object that is represented by the specified character
    \item \textit{add (char)}: the entity adds an instance of the specific character's class to the fortress
    \item \textit{transform (char)}: the entity will change classes altogether to an entirely different entity class (including those with different FSM definitions)
\end{itemize}


Each agent is initialized with the ``idle'' node as the starting base node, then a random subset of the possible action nodes is chosen from the simulation's provided configuration file to construct the rest of the graph. Each node can only be selected and added to the graph once to avoid redundancy and prevent over-complicated graph structures. 

Edges of the agent's FSM definition are conditions that can occur within the simulation to allow node transitions of the agent from state to state. Like the nodes, these conditions are also pre-programmed. In the case that a node has multiple edges, the conditions are ordered by priority within the system and the edge with the highest conditional priority transitions to its endpoint state. For this system, the possible condition edges to connect the action-state nodes of the graphs are as follows in the order of least to greatest priority:
\begin{enumerate}
    \item \textit{none}: no condition is required to transition states
    \item \textit{step} (int): every x number of simulation ticks the edge is activated and the node transitions
    \item \textit{within} (char) (int): checks whether the entity is within a number of spaces from an instance of another entity with the target character
    \item \textit{nextTo} (char): checks whether the entity is within one space (north, south, east, or west) of another entity of the target character
    \item \textit{touch} (char): checks whether the entity is in the same space as another entity of the target character
\end{enumerate}

The edges and their conditions are directional in relation to the nodes (i.e. an edge from node 0 to node 1 may have a different condition and definition than an edge from node 1 to node 0.) The entity can have from 0 to $n\times (n-1)$ edges. Like the node generation of the FSM, the edge connection and conditions are randomly generated from the subset of possible conditions provided in the simulation's configuration file. 

At any time during the simulation, the entity is always in a state at one of the set nodes. At each timestep---or update within the fortress environment---each connection is evaluated to move to the connecting node state based on whether the conditions are met, in order of priority defined internally. The agent then performs the action at its new current node on the next timestep. 


\subsection{Fortress}

The ``fortress'' class of the Amorphous Fortress contains the environment where the simulation takes place and stores general information accessible to all of the entities in the fortress. The borders and size of the fortress are defined initially with a set character width $w$ and height $h$ to enclose the entities. On initialization, the fortress generates each entity class FSM for each character defined in the passed configuration file. This global dictionary of entity classes allows any instance of an entity to add or transform different entity instances even if none currently exist on the map. The fortress maintains a list of currently active entity instances in the simulation and adds or removes them if called by ID value from an entity instance. The fortress also maintains positional data about each entity to return for conditional checks (i.e. whether a particular position has an entity located there.) The fortress also keeps a log of actions taken by each entity during simulation, and the timesteps at which they occurred. This log is exported upon the termination of the simulation by the engine. The fortress has 3 termination functions that are checked by the engine defined as follows:
\begin{itemize}
    \item \textit{extinction}: checks whether there are no entities at all left in the simulation
    \item \textit{overpopulation}: checks whether there exist more than $w\times h$ entities in the simulation
    \item \textit{inactivity}: checks whether an action has been taken by an entity in the last $x$ timesteps (for some integer $x$); achieved by checking the timestep value of the last logged action
\end{itemize}
The cause for termination is added to the end of the log, along with every entity class definition's FSM tree, and labeled by the seed used to generate the simulation for reproducibility.


\begin{figure}
    \centering
    \includegraphics[width=\linewidth]{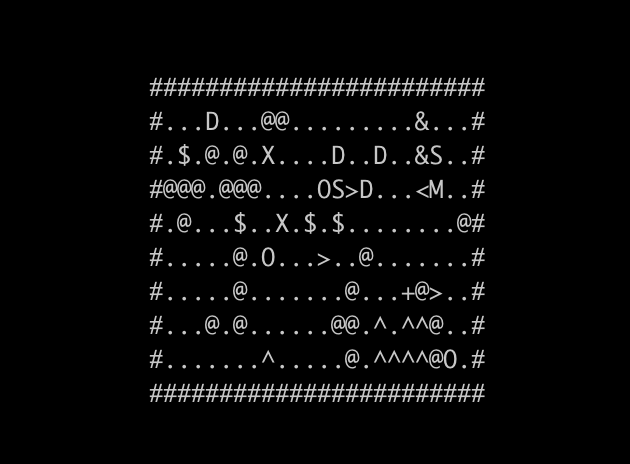}
    \caption{Sample fortress generated at random in the Amorphous Fortress interface. There are many instances of the entity classes predefined inside the fortress by the engine---determined by the configuration file}
    \label{fig:FortressExample}
\end{figure}

\subsection{Engine and Main Loop}

The engine serves as the entry point to the system and maintains the entire simulation as well as the update loop for the entities. The engine also imports the configuration file to pass to the fortress and to the entities - inclduing the seed value for reproducible randomization. The engine randomly selects from the character class definition to create new instances to add the fortress at random positions. Once the entities and their FSMs are populated in the Fortress, the engine starts and updates the simulation. Each entity is evaluated at its current node and any possible transitions from the said node. The engine continues updating the simulation until any of the termination conditions have been reached.

\subsection{Configuration File}

A configuration file that defines the action space and conditions can be provided to the system during initialization for the purpose of reproducibility. The parameters that can be specified in the configuration file are as follows:

\begin{itemize}
    \item \textit{seed (char)} - the seed number to specify a particular seed for reproducing that particular simulation or 'any' to create a new simulation
    \item \textit{character (str arr)} - the possible set of ASCII characters to represent the entities 
    \item \textit{action\_space (str arr)} - the possible set of action/nodes the entity can be initialized with for its FSM graph
    \item \textit{edge\_conditions (str arr)} - the possible set of conditions for a node to transition into another node for its FSM graph
    \item \textit{step\_range (int, int)} - the range of timesteps to randomly initialize the \textit{step} edge condition
    \item \textit{prox\_range (int, int)} - the range of spaces to randomly initialize the \textit{within} edge condition
    \item \textit{save\_log (boolean)} - `True' if the logs for the simulation should be saved, `False' if not 
    \item \textit{log\_file (char)} - the path to the file where the logs can be saved
    \item \textit{min\_log (int)} - the minimum number of steps for the simulation to have to save the log
    \item \textit{inactive\_limit (int)} - the number of seconds to let the simulation run before stopping due to inactivity
    \item \textit{pop\_perc (float)} - the chance for a copy of the entity's class to be added to the population
\end{itemize}

\subsection{Example Scenario}


\begin{figure}[t]
\centering
\begin{subfigure}[t]{.49\linewidth}
\centering
\includegraphics[width=.7\linewidth]{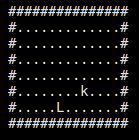}
\end{subfigure}
\begin{subfigure}[t]{.49\linewidth}
\centering
\includegraphics[width=.745\linewidth]{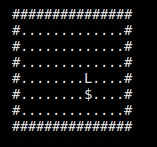}
\end{subfigure}
\caption{The example `Link' environment. Initially, Link (`L') moves randomly, and the Korok (`k') remains idle (left image.) Once Link is adjacent to the Korok, it transforms into a seed (`\$'), at which point Link pursues the seed and removes it from the map (right image.)}
\end{figure}

To illustrate the potential of our system to generate plausible simulation dynamics, we use our domain-specific language to define a simple example environment by hand. The `Zelda' environment, inspired by the recent franchise entry \textit{The Legend of Zelda: Tears of the Kingdom}~\citep{zelda_tok}, contains two initial entities: `Link' and `Korok'. In the Zelda franchise, Koroks are entities normally revealed after the player, controlling the character Link, completes a puzzle or interacts with a unique hidden object, at which point the Korok gifts the player a seed that can be used as in-game currency. 

In our toy environment, a single Link entity and a single Korok entity are spawned at random positions on the map. Link's FSM (\autoref{fig:EntityFSM}) is defined such that the character will move randomly on the map at every other timestep. When the Korok (`k') is directly adjacent to Link (`L'), it transforms into `\$' (intended to represent a Korok seed). When Link is within $3$ tiles of a `\$' seed, he will begin to chase it, and when he is overlapping with it, he takes the seed, removing the `\$' character from the map. (The `\$' character's FSM has only a root node and therefore remains idle until it is removed from the map by Link.)

\section{Evolutionary Experiment}

To explore the potential depths of the generated finite state machine behaviors, we implement a hillclimbing evolutionary algorithm to evolve the agents and fortress towards having the largest, but also most traversed FSMs. The mutations of this algorithm were applied at the global character definition class level - such that each future instance of an entity class would all have the same modified FSM tree definition. The number of instances and types of entities instantiated onto the fortress were also mutated from generation to generation. We chose to use the Hillclimber evolutionary algorithm for this study as we were only concerned with exploring how the fortress and simulation could gradually evolve toward an objective as a whole. Using the hillclimber algorithm for this experiment, also allowed for the simplest open-ended emergence for novel entity behavior patterns. The motivation for these experiments were to see if ``interesting'' behaviors could emerge from such a simple evolutionary method in this constrained environment. Future work would look to examine the diversity of behaviors exhibited by the agents - most likely using a novelty search \citep{lehman2008exploiting} or with a quality diversity algorithm such as MAP-Elites \citep{mouret2015illuminating}.

Three different mutation algorithms were implemented for evolution. The first involved mutating the nodes of randomly selected entity classes. This was done by either 1) deleting random nodes (so long as there is more than one node in the graph) 2) adding more nodes (if there are any that have not been added already) or 3) replacing nodes with another action. Similarly, the second mutation modified the edges by either 1) deleting random edges (so long as there was more than one edge in the graph), 2) adding edges between nodes, or 3) overwriting the conditional check on an edge. The third mutation modified the number of instances in the fortress by adding or removing instances. From generation to generation, only the ASCII character representation and the position of the entity are stored, therefore each new generation has a distinctly new instance created for the simulation. After mutation, the modified entity class FSM was pruned up for any unconnected edges and orphaned nodes after mutation of the finite state machines. The mutation was done by coin-flip chance, given some pre-set probability for both the node level, edge level, and instance level mutation if a random number was within the value for probability a mutation was performed. This continued until the random value reached above the threshold of the probability. Algorithm \ref{alg:mutation} demonstrates the mutation process for the evolutionary system of the fortress.  

\begin{algorithm}[ht!]
\caption{Mutation function for the Fortress}\label{alg:mutation}
\KwInput{$node\_prob$, $edge\_prob$, $instance\_prob$}

$node\_r$ = random()\;
$edge\_r$ = random()\;
$instance\_r$ = random()\;

\tcc{Mutate random entity class nodes}
\While{$node\_r < node\_prob$}{
    $i$ = random(0,2)\;
    $e$ = random($fortress.ent\_def$)\;
    \uIf{$i == 0$}{
        $fortress.\_delete\_node(e)$\;
    }\uElseIf{$i == 1$}{
        $fortress.\_add\_node(e)$\;
    }\uElseIf{$i == 2$}{
        $fortress.\_alter\_node(e)$\;
    }
    $node\_r$ = random()\;
}
\tcc{Mutate random entity class edges}
\While{$edge\_r < edge\_prob$}{
    $i$ = random(0,2)\;
    $e$ = random($fortress.ent\_def$)\;
    \uIf{$i == 0$}{
        $fortress.\_delete\_edge(e)$\;
    }\uElseIf{$i == 1$}{
        $fortress.\_add\_edge(e)$\;
    }\uElseIf{$i == 2$}{
        $fortress.\_alter\_edge(e)$\;
    }
    $edge\_r$ = random()\;
}
\tcc{Mutate random entity instances in the fortress}
\While{$instance\_r < instance\_prob$}{
    $i$ = random(0,1)\; 
    $e$ = random($fortress.entities$)\;
    \uIf{$i == 0$}{
        $fortress.\_remove\_entity(e)$\;
    }\uElseIf{$i == 1$}{
        $x,y$ = random(fortress.pos)\;
        $fortress.\_add\_entity(e,x,y)$\;
    }
    $instance\_r$ = random()\;
}

\end{algorithm}


\begin{figure}
    \centering
    \includegraphics[width=\linewidth]{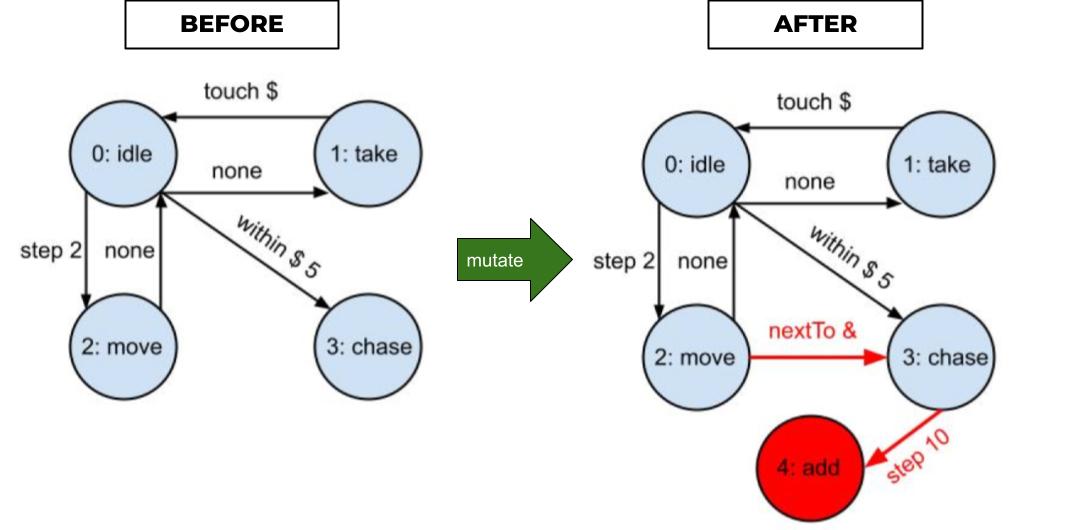}
    \caption{Example of the FSM being mutated. A node and 2 edges are both added in this modification.}
    \label{fig:MutationExample}
\end{figure}

In this experiment, we wanted to encourage more diverse and deep behaviors from the generated FSMs of the fortress. Therefore, our fitness function had to be based on if a node or edge of an entity's graph was traversed during the entire course of the simulation. The fitness function for evolving the fortresses using the hill-climbing algorithm was defined as the following equation: 

\begin{equation}
fitness = \frac{v}{u+1} \times t
\end{equation}

where \textit{v} is the sum of the number of traversed (or visited) nodes and edges of every entity class FSM in the fortress, \textit{u} is the combined number of unused (or unvisited) nodes and edges of each entity class FSM, and \textit{t} as the total number of nodes and edges in the entity class FSMs. The visitation of a node or edge in an entity class's FSM was determined by the combined behaviors of every instance of a particular entity. Because some instances of an entity class may have different experiences and interactions based on initial placement, we used the global definition of the combined behaviors for the fitness function. For example, if a single instance of the `@' class completes an action that is part of the FSM's tree due to a conditional check successfully occurring, both the node it traveled to and the edge condition it traveled by would be considered 'visited' for the entire entity class of `@'. 
This fitness function was also defined to encourage both the generation of large but thoroughly explored FSMs for the fortress overall and not just by a single entity class.


For this study, we examine the evolution of 5 fortresses each starting from randomly generated seeds. We evolved the fortress for a total of 1000 generations -- comparing the fitness of one new mutated fortress to the best-generated fortress. If the generated fortress achieved a higher fitness score, the best fortress would be replaced by the current instance, and so on. The fortress was simulated for 20 timesteps before being evaluated for entity tree traversals. This was to limit the computational calculations of each instance in the fortress while still allowing for depth and interaction with the agent behaviors. This study used a $50\%$ independent probability for all mutation chances described before -- a $50\%$ chance to mutate a randomly selected entity class's FSM nodes, edges, and/or to add or remove an instance from the fortress. A total of $15$ unique ASCII characters were used for the experiments, and all edge conditions and node action conditions were allowed for selection. 

\section{Results and Discussion}

\begin{figure}
    \centering
    \includegraphics[width=\linewidth]{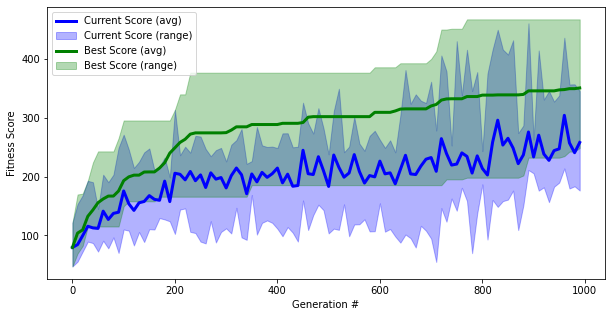}
    \caption{Best and current fitness scores of the 5 experiment trials (averaged and min/max range}
    \label{fig:score_graph}
\end{figure}

We observed a number of relatively interesting behaviors from both the evolutionary experiment, the final entity FSM definitions, the fortress construction, and the trends of the fortress evolution. Figure \ref{fig:score_graph} shows the historical statistical trend of the fortress composition over the 1000 generations over the 5 trials; the fitness score of the best-saved fortress, the comparison score of the mutated fortress, and the number of entities found in the mutated fortress. The following subsections describe these phenomena in more detail along with our insights and hypothesis for these emergent behaviors.

\subsubsection{Upper bound fitness}
The fortress environment is able to successfully improve its fitness score for maximizing coverage and size of trees for each entity definition in the fortress via simulation. Figure \ref{fig:score_graph} demonstrates gradual improvement of the fortress's fitness score with increased generations across all 5 trials. Due to different starting instances and behaviors, the final score is not consistent over all 5 seeds, however, the mean final score was $345.2$ with a standard deviation of $60.97$. The maximum potential calculated score is $19440$ -- given a fully connected graph with edges between each pair of distinct nodes in the FSM for each character in the defined configuration; $max\_score = (n\times(n-1)+n)\times c$ where $n$ is the number of node types and $c$ is the number of characters available. However, having every entity with a maximally large FSM graph may not lead to very interesting behavior, since in such a case, every entity type would have effectively identical FSMs and behavior profiles. 
The fitness score of offspring is extremely variable during evolution as the best fortress score increased, indicating that the fitness landscape is highly non-local given our mutation operators and that small mutations can lead to vastly different simulation dynamics. The trend line of all of the experiments for the most part consistently follows the best fortress score. Towards the later generations, the line starts to level out and stay stagnant.

\subsubsection{Number of Entities}

\begin{figure}
    \centering
    \includegraphics[width=\linewidth]{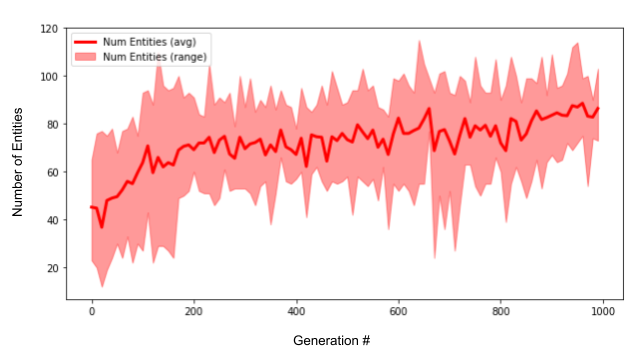}
    \caption{Number of entities in the current fortress of the 5 experiment trials (averaged and min/max range}
    \label{fig:entity_graph}
\end{figure}

With the defined fitness function, the environment tends to favor making as many entities as possible in the fortress. Overall, the $5$ best fortresses tend to max out the number of entities possible before reaching an ``overpopulation'' termination condition defined in the system. 
One possible explanation for this tendency is that as the number of repeated instances of each entity on the map increases, it becomes more likely that different instances will visit distinct regions of their FSM tree over the course of the simulation, thus contributing to the proportion of visited nodes and edges.
Were the fortress size to be increased beyond the currently defined area of $78$ spaces ($13$ characters wide, $6$ characters tall, excluding borders) evolution might show even faster progress toward higher fitness scores. Figure \ref{fig:entity_graph} shows the trend of the number of entities for all 5 experiment trials.

\subsubsection{FSM Objective Coverage}
Globally within the fortress, each entity from the best fortress of each trial successfully optimized towards the fitness function's objective. A large majority of the entity classes in the final best fortresses had high graph traversal coverage overall. The calculated average across all 5 fortresses for every entity class had an average of 72\% node coverage and 58\% edge coverage. The function was designed with the intention to reward large, well-traversed finite state machine graphs - while giving fewer rewards to large graphs with less traversal and small graphs with fuller traversal. While many generated entity classes had large graphs - defined as having 7 or 8 nodes and between a range of 15-30 edges - and strong node/edge coverage, there was also a small number of entities that had significantly smaller graphs - with 2 or fewer nodes and less than 5 edges. This diversity of entities regardless of the size of the graphs may have been due to a naturally emerging ``symbiosis'' of the entities and their relationships to each. For example, although one entity class may not have a more actively influential node such as ``take'' or ``add'', the conditional edges of another entity's class could have depended on the presence of that entity in order to do something more influential - i.e. another entity may have needed to be ``nextTo'' or ``touch''ing the small-graph entity in order to ``clone'' itself or ``transform''.

\subsubsection{Rainbow Goop FSM Generation}

\begin{figure}
    \centering
    \includegraphics[width=\linewidth]{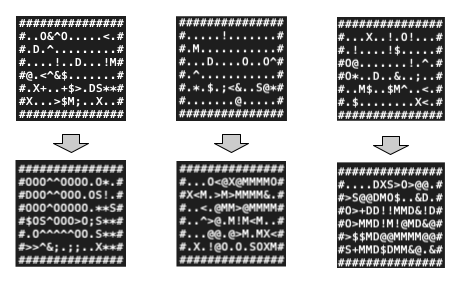}
    \caption{Initial and final renderings of the fortress for 3 seeds}
    \label{fig:init_final_render}
\end{figure}

Because of the objective function, the FSM shifted towards having ``reproductive'' behaviors. The entities evolved towards focusing on having node actions that created more of themselves but also changed themselves into another entity. As such, the ``add', ``transform'', and ``clone'' nodes were seen more frequently in the entities graphs than more destructive nodes such as ``take'' and ``die.'' Based on the initial and final map renderings, many of the initial entities on the map would try to clone itself to create more copies of itself - this allowed for deeper graph exploration to fulfill the ``visited'' over ``unvisited'' node statistic. Afterward, many entities would often transform into other entities - to fulfill other missing nodes and edge visitations concerning other entities that may not have initially been on the map. This explains the copious - almost overpopulated - amount of entities in the final rendering of the fortress simulation. Figure \ref{fig:init_final_render} shows the initial fortress renderings and the final fortress renderings after 20 steps of simulation. More diversity of characters and entities on the map also encouraged more interactions overall. We call this type of behavior ``rainbow goop'' as the behavior replicates itself, but also tries to transform itself into other entities to diversify the fortress.

\section{Conclusion}

We introduce an abstract life simulation to observe emergent AI behaviors generated based on finite-state machines. We evaluate and explore this simulation's potential via the Hillclimbing algorithm and find many unexpected but cool interactions given such a limited system.

Future work could also include a more sophisticated fitness function. In particular, it would be interesting to explore the measure of complexity, either as entropy, compressibility, predictability, or any related operationalization of a complexity concept. There being many types of complexity, one could empirically investigate which one corresponds to human notions of interestingness. For example, the complexity of instantaneous game states is probably less interesting than the complexity of trajectories over time.

One might also try to find games that are learnable, similar to the fitness function of~\cite{togelius2008experiment}. Converging theory from machine learning, game design, and developmental psychology suggests that learnable tasks are more interesting. This would require introducing a player interface to the game. This system overall offers many opportunities for exploration with emergent AI and we look forward to future developments within this domain.

\section{Acknowledgements}
We thank Noelle Law for help with generating these amazing graphs for this paper and the emotional support while we wrote this paper in the late hours of the night.

\footnotesize
\bibliographystyle{apalike}
\bibliography{ref} 

\end{document}